\RequirePackage{fix-cm}
\documentclass[smallextended]{svjour3}       
\smartqed  
\usepackage{lineno,hyperref}
\usepackage{times}
\usepackage{epsfig}
\usepackage{graphicx}
\usepackage{amsmath}
\usepackage{amssymb}
\usepackage{multirow}
\usepackage{caption}
\usepackage{subcaption}
\usepackage{float}
\usepackage{appendix}
\usepackage{color}

\modulolinenumbers[5]

\begin{document}

\title{
Lipschitz Constrained GANs via Boundedness and Continuity	
}


\author{Kanglin Liu         \and
        Guoping Qiu 
}


\institute{Kanglin Liu$^{1,2,3}$ \at                            
              \email{max.liu.426@gmail.com}           
           \and
           Guoping Qiu$^{1,2,3,4}$\at
              \email{Guoping.Qiu@nottingham.ac.uk} \\             
              1.Shenzhen University, Shenzhen, China\\
              2.Guangdong Key Laboratory of Intelligent Information Processing, Shenzhen, China\\
              3.Shenzhen Institute of Artificial  Intelligence and Robotics for Society, Shenzhen, China\\
              4.University of Nottingham, Nottingham, United Kingdom          
}

\date{Received: date / Accepted: date}

\maketitle

\begin{abstract}

One of the challenges in the study of Generative Adversarial Networks (GANs) is the difficulty of its performance control. Lipschitz constraint is essential in guaranteeing training stability for GANs. Although heuristic methods such as weight clipping, gradient penalty and spectral normalization have been proposed to enforce Lipschitz constraint, it is still difficult to achieve a solution that is both practically effective and theoretically provably satisfying a Lipschitz constraint. In this paper, we introduce the boundedness and continuity ($BC$) conditions to enforce the Lipschitz constraint on the discriminator functions of GANs. We prove theoretically that GANs with discriminators meeting the BC conditions satisfy the Lipschitz constraint. We present a practically very effective implementation of a GAN based on a convolutional neural network (CNN) by forcing the CNN to satisfy the $BC$ conditions (BC-GAN). We show that as compared to recent techniques including gradient penalty and spectral normalization, BC-GANs not only have better performances but also lower computational complexity.

\end{abstract}

\section{Introduction}
\label{intro}

Generative Adversarial Networks (GANs) \cite{A1} is hailed as one of the most significant developments in machine learning research of the past decade. Since its first introduction, GANs have been applied to a wide range of problems and numerous papers have been published. In a nutshell, GANs are constructed around two functions \cite{A5,A6}: the generator $G$, which maps a sample $z$ to the data distribution, and the discriminator $D$, which is trained to distinguish real samples of a dataset from fake samples produced by the generator. With the goal of reducing the difference between the distributions of fake and real samples, a GAN training algorithm trains $G$ and $D$ in tandem. 

A major challenge of GANs is that controlling the performance of the discriminator is particularly difficult. Kellback-Leibler (KL) divergence was originally used as the loss function of the discriminator to determine the difference between the model and target distributions \cite{A2}. However, KL divergence is potentially non-continuous with respect to the parameters of $G$, leading to the difficulty in training \cite{A3,A4}. Specifically, when the support of the model distribution and the support of the target distribution are disjoint, there exists a discriminator that can perfectly distinguish the model distribution from that of the target. Once such a discriminator is found, zero gradients would be back propagated to $G$ and the training of  $G$ would come to a complete stop before obtaining the optimal results. Such a phenomena is referred to as the vanishing gradient problem.

The conventional form of Lipschitz constraint  is given by: $||f(x_1)-f(x
_2)||\le k\cdot||x_1-x_2||$. It is obvious that Lipschitz constraint requires the continuity of the constrained function and guarantees the boundedness of the gradient norm. Besides, it has been found that enforcing Lipschitz constraint can provide provable robustness against adversarial examples \cite{A24}, improve generalization bounds \cite{A25},  enable Wasserstein distance estimation \cite{A8}, and also alleviate the training difficulty in GANs. 
Thus, a number of works have advocated the Lipschitz constraint.
To be specific, weight clipping was first introduced to enforce the Lipschitz constraint \cite{A3}. However, it has been found that weight clipping may lead to the capacity underuse problem where training favors a discriminator that uses only a few features \cite{A8}. To overcome the weakness of weight clipping, regularization terms like gradient penalty are added to the loss function to enforce Lipschitz constraint on $D$ \cite{A8,A9,A26}. 
More recently, Miyato \textit{et al.} \cite{A10} introduce spectral normalization to control the Lipschitz constraint of  $D$ by normalizing the weight matrix of the layers, which is regarded as an improvement on orthonormal regularization \cite{A11}.
Using gradient penalty or spectral normalization can stabilize the training and gain improved performance. However, it has been found that gradient penalty suffers from the problem of not being able to regularize the function at the points outside of the support of the current generative distribution  \cite{A10}. 
In addition, spectral normalization has been found to suffer from the problem of gradient norm attenuation \cite{A22,A23}, \textit{i.e.}, a layer with a Lipschitz bound of 1 can reduce the norm of the gradient during backpropagation, and each step of backprop gradually attenuates the gradient norm, resulting in a much smaller Jacobian for the network’s function than is theoretically allowed.
Also as we will show in Section \ref{sec3} and Section \ref{sec4.4}, these new methods have the capacity underuse problem (see Proposition 1 and Figure \ref{fig:fig1} ). Therefore, despite recent progress, it remains challenging to achieve  practical success
as well as provably satisfying a Lipschitz constraint.

In this paper, we introduce the boundedness and continuity ($BC$) conditions to enforce the Lipschitz constraint, and introduce a CNN based implementation of GANs with discriminators satisfying the $BC$ conditions.
We make the following contributions:

(a) We prove that SN-GANs, one of the latest GAN training algorithms that use spectral normalization, will prevent the discriminator functions from obtaining the optimal solution when applying Wasserstein distance as the loss metric even though the Lipschitz constraint is satisfied. 

(b) We present $BC$ conditions to enforce the Lipschitz constraint for the GANs' discriminator functions, and introduce a CNN based implementation of GANs by enforcing the $BC$ conditions (BC-GANs). We show that the performances of BC-GANs are competitive to state of the art algorithms such as SN-GAN and WGAN-GP but having  lower computational complexity.   


\section{Related Work}
\subsection{Generative Adversarial Networks (GANs)}
Generative adversarial networks (GANs) is a special generative model to learn a generator $G$ to capture the data distribution via an adversarial process. Specifically, a discriminator $D$ is introduced to distinguish the generated images from the real ones, while the generator $G$ is updated to confuse the discriminator. The adversarial process is formulated as a minimax game as:
\begin{equation} \label{key1}
\underset{G}{\mathrm{min}} \ \underset{D}{\mathrm{max}} V(G, D)
\end{equation}
where min and max of $G$ and $D$ are taken over the set of the generator and discriminator functions respectively. $V(G, D)$ is to evaluate the difference in the two distributions of $q_x$ and $q_g$, where $q_x$ is the data distribution, and $q_g$ is the generated distribution.
The conventional form of $V(G, D)$ is given by Kellback-Leibler (KL) divergence: $ E_{x \sim q_{x}}[\mathrm{log}D(x)]+E_{x'\sim q_{g}}[\mathrm{log}(1-D(x'))]$ \cite{A2}.

\subsection{Methods to Enforce Lipschitz Constraint}

Applying KL divergence as the implementation of $V(G, D)$  could lead to the training difficulty, \textit{e.g.}, the vanishing gradient problem. Thus,  numerous methods have been introduced to solve this problem by enforcing the Lipschitz constraint, including weight clipping \cite{A3}, gradient penalty \cite{A5} and spectral normalization\cite{A10}. 

Weight clipping was introduced by Wasserstein GAN (WGAN) \cite{A3}, which used Wasserstein distance to measure the differences between real and fake distributions instead of KL divergence.
\begin{equation}\label{key2}
W(P_r, P_g) = \mathop {\sup }\limits_{f \in Lip1} \mathop E\limits_{x \sim P_r} [f(x)] - \mathop E\limits_{x \sim P_g} [f(x)]
\end{equation}
where $W(P_r, P_g)$ represents the Wasserstein distance,  $P_r$ and $P_g$ are the real and fake distributions, respectively. Weight clipping enforces the Lipschitz constraint by truncating each element of the weight matrices. 
Wasserstein distance shows superiority over KL divergence, because it can effectively  avoid the vanishing gradient problem brought by KL divergence.
In contrast to weight clipping, gradient penalty \cite{A8} penalizes the gradient at sample points to enforce Lipschitz constraint:
\begin{equation}\label{key3}
{L_D} = E[f(G(z))]{\rm{ }} - E[f(x)]{\rm{ + }}\underbrace {\alpha E[{{(||\nabla f(x)|| - 1)}^2}]}_{{\rm{gradient \ penalty}}}
\end{equation}
where $L_D$ is the loss objective  for the discriminator, and $\alpha$ is a hyperparameter. 

Spectral normalization is a weight normalization method, which controls the Lipschitz constraint of the discriminator function by literally constraining the spectral norm of each layer.
The implementation of the spectral normalization can be expressed as:
\begin{equation}\label{key4}
W_{SN}(W): = W/\sigma (W)
\end{equation}
where $W$ represents the weight matrix in each network layer, $\sigma (W)$ is the spectral norm of matrix $W$, which equals to the largest singular value of the matrix $W$, and $W_{SN}(W)$ represents the normalized weight matrix.  To a certain extent, spectral normalization have succeeded in facilitating stable training and improving performance. 
\section{Existing Problems}\label{sec3}
Although heuristic methods have been proposed to enforce Lipschitz constraint, it is still difficult to achieve a solution that is both practically effective and theoretically provably satisfying the Lipschitz constraint. To be specific,  weight clipping was proven to be unsatisfactory in \cite{A5}, and it can lead to the capacity underuse problem where training favors a discriminator that uses only a few features \cite{A8}. In addition,
gradient penalty suffers from the obvious  problem of not being able to regularize the function at the points outside of the support of the current generative distribution. In fact, the generative distribution and its support gradually changes in the course of the training, and this can destabilize the effect of the regularization itself \cite{A10}. 
Moreover, it has been found that spectral normalization suffers from the gradient norm attenuation problem \cite{A22,A23}.  Furthermore, we have found that applying spectral normalization prevents the discriminator functions from obtaining the optimal solutions when using Wasserstein distance as the loss metric. To provide an explanation to this problem, we present Proposition 1.

Let $P_r$ and $P_g$ be the distributions of real images and generated images in $X$, a compact metric space. The discriminator function $f$ is constructed based on a neural network of the following form with input $x$:
\begin{equation} \label{key5}
f(x,\theta )=W^{L+1}a_L(W^L(a_{L-1}(\cdots a_1(W^1x)))))
\end{equation}	
where $\theta :=\{ W^1, W^2, ..., W^{L+1} \}$ is the learning parameter set, and $a_l$ is an element-wise non-linear activation function.  Spectral normalization is applied on $f$ to guarantee the Lipschitz constraint.  

{\noindent \bf Proposition 1} When using Wasserstein distance as the loss metric of $f$, the optimal solution to $f$  is unreachable.

\section{Enforcing Boundedness and Continuity in CNN based  GANs }
Finding a proper way to enforce the Lipschitz constraint remains an open problem. Motivated by this, we search for a better way to enforce the Lipschitz constraint.

\subsection{ $BC$ Conditions}

The purpose is to find the discriminator from the set of $k$-Lipschitz continuous functions \cite{A12}, which obeys the following condition:
\begin{equation}\label{key6}
||f({x_1}) - f({x_2})|| \le k||{x_1} - {x_2}||
\end{equation}

Equation (\ref{key6}) is referred to as the Lipschitz continuity or Lipschitz constraint. If the discriminator function $f$ satisfies following conditions, it is guaranteed to meet the condition of Equation (\ref{key6}) :

(a) Boundedness: $f$ is a bounded function.

(b) Continuity: $f$ is a continuous function, and the number of points where $f$ is continuous but not differentiable is finite. Besides, if $f$ is differentiable at point $x$, its derivative is finite.

Conditions (a) and (b) are referred to as the boundedness and continuity ($BC$) conditions. A discriminator satisfying the $BC$ conditions is referred as a Bounded Discriminator, and a GAN model with $BC$ conditions enforced is referred to as BC-GAN. Following Theorem 1 and Theorem 2 guarantee that meeting the $BC$ conditions is sufficient to enforce the Lipschitz constraint of Equation (\ref{key6}). (see proofs in Appendix)

{\noindent \bf Theorem 1.} Let $\Psi$ be the set of all $f: X  \rightarrow R $, where $f$ is a continuous  function. In addition, the number of points where $f$ is continuous but not differentiable is finite. Besides, if $f$ is differentiable at point $x$, its derivative is finite. Then, $f$ in $\Psi$ satisfies Lipschitz constraint.

{\noindent \bf Theorem 2.}  Let $P_r$ and $P_g$ be the distributions of real images and generated images in $X$, a compact metric space. Let $\Omega$ be the set of all $f: X  \rightarrow R$, where $f$ is a continuous and bounded  function. And, the number of points where $f$ is continuous but not differentiable is finite. Besides, if $f$ is differentiable at point $x$, its derivative is finite. The set  $\Omega$ can be expressed as:
\begin{equation}\label{key7}
\Omega:{\rm{\{ }}f |  {\rm{ }}||f(x)|| \le m,{\rm{ if }}\frac{{\partial f(x)}}{{\partial x}}{\rm{exists, }}||\frac{{\partial f(x)}}{{\partial x}}|| < \infty {\rm{\} }}
\end{equation}
where $m$ represents the bound. Then, there must exist a $k$, and we have a computable $k \cdot W(P_r, P_g)$:
\begin{equation}\label{key8}
k \cdot W({P_r},{P_g}) = \mathop {\sup }\limits_{f \in \Omega } \mathop E\limits_{x \sim {P_r}} [f(x)] - \mathop E\limits_{x \sim {P_g}} [f(x)]
\end{equation}
where $W(P_r, P_g)$ represents the Wasserstein distance  between $P_r$ and $P_g$ \cite{A1,A4}.

According to Theorem 1  and Theorem 2 , it is obvious that the $BC$ conditions are sufficient to enforce the Lipschitz constraint. Furthermore, $k \cdot W(P_r, P_g)$ is bounded and computable, and can be obtained as:
\begin{equation}\label{key9}
k \cdot W({P_r},{P_g}) = \mathop {\max }\limits_{f \in \Omega } \mathop E\limits_{x \sim {P_r}} [f(x)] - \mathop E\limits_{_{z\sim{p(z)}}} [f(G(z))]
\end{equation}

Then, $k \cdot W(P_r, P_g)$ can be applied as a new loss metric to guide the training of $D$. Logically, the new objective for  $D$ is:
\begin{equation}\label{key10}
{L_D}{\rm{ = }}\mathop {\min }\limits_{f \in \Omega } {E_{z\sim{p(z)}}}[f(G(z))] - {E_{x\sim{{P_r}}}}[f(x)]
\end{equation}

Theorem 3 in \cite{A3} tells us that,
\begin{equation}\label{key11}
{\nabla _\theta }kW({P_r},{P_g}){\rm{ = }} - {E_{z\sim{p(z)}}}[{\nabla _\theta }f(G(z))]
\end{equation}
where $\theta$ is the parameters of $G$. Equation (\ref{key11}) indicates that using gradient descent to update the parameters in $G$ is a principled method to train the network of $G$. Finally, the new objective for  $G$ can be obtained:
\begin{equation} \label{key12}
{L_G}{\rm{ = }}\mathop {\min }\limits_\theta  - {E_{z\sim{p(z)}}}[f(G(z))]
\end{equation}

\subsection{Implementation of $BC$ Conditions}
In this paper, we introduce a simple but efficient implementation of $BC$ conditions. 
When applying the $BC$  conditions to $D$, the training of $D$ can be equivalently regarded as a conditional (constrained) optimization process. Then, Equation (\ref{key10}) can be updated as:
\begin{equation}\label{key13}
\begin{split}
&\mathop {\min }\limits_{f \in \Omega } \{ {E_{z\sim{p(z)}}}[f(G(z))] - {E_{x\sim{{P_r}}}}[f(x)]\} \\
&s.t.{\rm{  ||}}f(x)|| \le m,{\rm{  if }}\frac{{\partial f(x)}}{{\partial x}}{\rm{exists, }}||\frac{{\partial f(x)}}{{\partial x}}|| < \infty 
\end{split}
\end{equation}

\begin{table*}[htp]
	\centering
	\begin{tabular}{l}
		\hline \hline
		{\bf Algorithm 1: BC-GAN} \\ 
		\hline 
		Require:\\
		the number of $D$ iteration per $G$ iteration \textit{n\textsubscript{critic}},\\
		the batch size  \textit{n}, the bound \textit{m}, 	\\
		initial critic parameter \textit{w}\textsubscript{0}, \\
		initial generator parameters $\theta$\textsubscript{0}\\
		1: {\bf while} $\theta$ has not converged {\bf do}\\
		2: \ \ \ \ \ \ Sample $\{ {x^{(i)}}\} _{i = 1}^n$ $\sim$ \textit{P\textsubscript{r}}\\
		3: \ \ \ \ \ \ Sample $\{ {z^{(i)}}\} _{i = 1}^n$ $\sim$ \textit{P\textsubscript{z}}\\
		4: \ \ \ \ \ \ {\bf for} \textit{t}=1,2,...,\textit{n\textsubscript{critic}} {\bf do}\\
		5: \ \ \ \  \ \ \ \ \ \ $\frac{1}{n}\sum\nolimits_{i = 1}^n {f({x^i})} $$\rightarrow$\textit{L\textsubscript{r}}\\
		6: \ \ \ \  \ \ \ \ \ \ $\frac{1}{n}\sum\nolimits_{i = 1}^n {f(g({z^i}))} $$\rightarrow$\textit{L\textsubscript{g}}\\
		7:\ \ \ \ \ \ \ \ \ \ \  $[{L_g} - {L_r} + \beta  \cdot {\rm{max(\|}}f(x){\rm{\|}} - m,0{\rm{)}}]$$\rightarrow$\textit{L\textsubscript{D}}\\
		8: \ \ \ \ \ \ \ \ \ \  Adam($\triangledown$\textit{\textsubscript{w}}\textit{L\textsubscript{D}})$\rightarrow$\textit{w}\\
		9: \ \ \ \ \ \ \ \ end for\\
		10:\ \ \ \ \ \ Adam(${\nabla _\theta }[-\frac{1}{n}\sum\nolimits_{i = 1}^n {f(g({z^i}))} ]$)$\rightarrow$$\theta$\\
		11: end while
		\\ 
		\hline \hline
	\end{tabular} 
\end{table*}

In this paper, the discriminator function $f$ is implemented by a deep neural network, which applies a series of convolutional  and non-linear operations. Both convolutional and non-linear functions are continuous, which means that $D$ is a continuous function.  Moreover, the gradients of the output of $D$ with respect to the input are always finite. As a result, condition (b) is satisfied naturally. 
To guarantee condition (a), the Lagrange Multiplier Method can be applied here, then the objective of $D$ can be written as the following equation:
\begin{equation}\label{key14}
\begin{split}
{L_D}{\rm{ = }}&\mathop {\min }\limits_f \{ {E_{z\sim{p(z)}}}[f(G(z))] - {E_{x\sim{{P_r}}}}[f(x)]\} \\
&+ \beta  \cdot {\rm{max(||}}[f(x)]{\rm{||}} - m,0{\rm{)}}
\end{split}
\end{equation}
where $\beta$ is the hyperparameter and $m$ represents the bound. The term $\rm{max}$$(\left \| f(x) \right \|-m, 0)$ plays the role of forcing $D$ to be a bounded function, while $E_{z \sim p(z)}\left [ f(G(z))\right ] - E_{x \sim p(x)}\left [ f(x) \right ]$ is  used to determine $k \cdot W(P_r, P_g)$.
The procedure of training the BC-GAN is described in Algorithm 1.

\subsection{Validity}\label{sec4.4}
In order to verify the validity of proposed $BC$ conditions, we use synthetic datasets as those presented in \cite{A9} to test discriminator's performance. Specifically, discriminators are trained to distinguish the fake distribution from the real one. The toy distributions hold the fake distribution $P_g$ as the real distribution $P_r$ plus unit-variance Gaussian noise. Theoretically, discriminator with good performance is more likely to learn the high moments of the data distributions and model the real distribution. 
Figure \ref{fig:fig1} illustrates the value surfaces of the discriminator. 
It is clearly seen that discriminator enforced by $BC$ conditions have a good performance on discriminating the real samples from the fake ones, demonstrating the validity of proposed method.

\subsection{Comparison with Spectral Normalization and Gradient Penalty}

Gradient penalty, spectral normalization as well as our proposed method are inspired by different motivations to enforce the Lipschitz constraint on $D$. Therefore, they differ in the way of implementation and in principle.
The first difference is the way of implementation. Gradient penalty and our method operate on the loss function directly, while spectral normalization constrains the weight matrix instead of the loss metric. 

Secondly, they differ in principle. For BC-GAN, $k \cdot W(P_r, P_g)$ is applied to evaluate the difference between the fake and real distributions instead of $W(P_r, P_g)$, which is used in WGAN-GP and WGAN. Moreover, WGAN-GP and SN-GAN strictly constrain the Lipschitz constant to be 1 or a  known constant. While BC-GAN eases the restriction on the Lipschitz constant, and $k$ is an unknown scalar parameter which will have no influence on the training of the network. Therefore, $k \cdot W(P_r, P_g)$ can be employed as a new loss metric to guide the training of $D$.

\begin{figure*}
	\centering
	\begin{center}
		\includegraphics[width=0.7\linewidth]{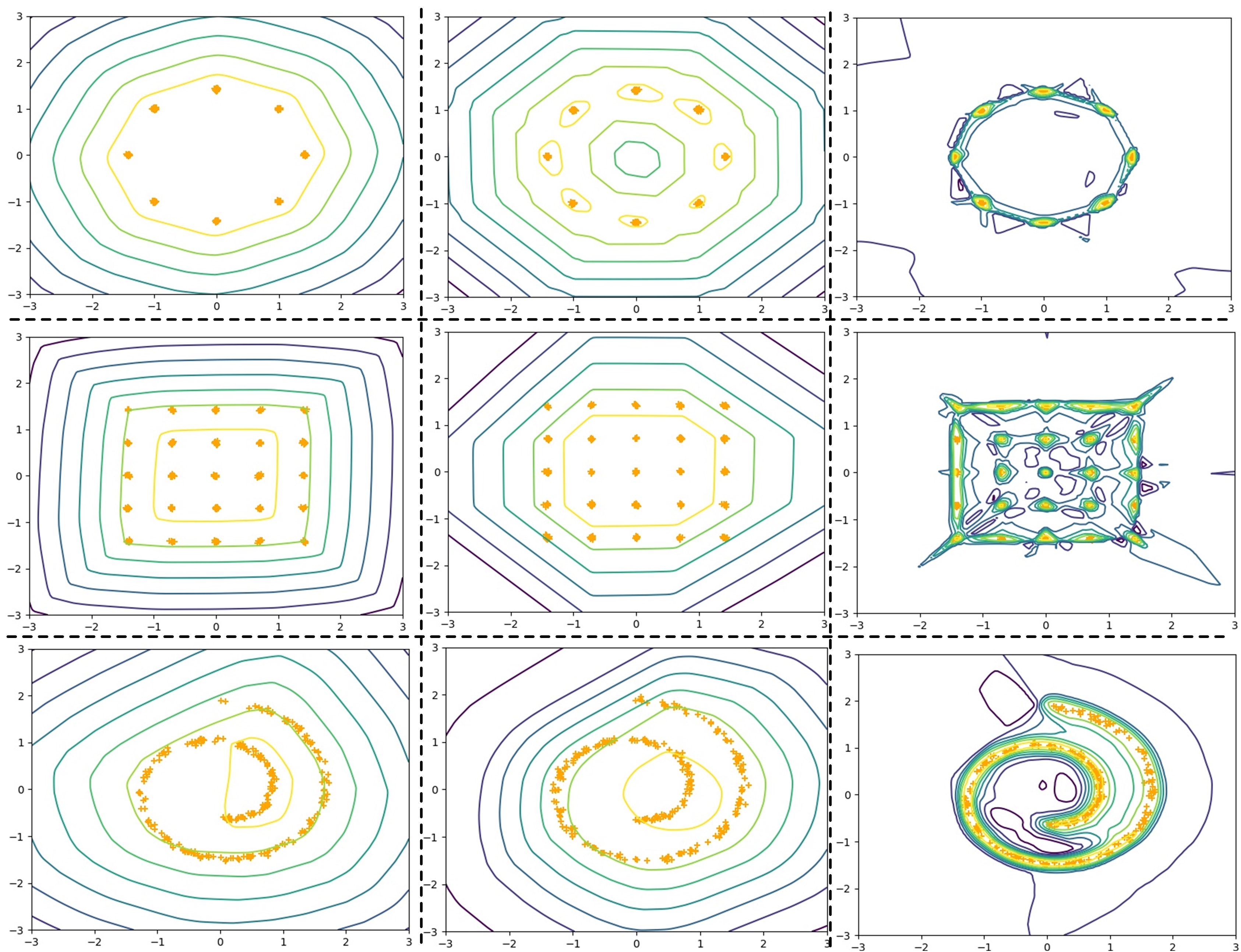}
	\end{center}
	\caption{Value surface of the discriminators trained to optimality on toy datasets. The yellow dots are data points, the lines are the value surfaces of the discriminators. Left column: Spectral Normalization. Middle column: Gradient Penalty. Right column: The proposed method. The upper, middle and lower rows are trained on 8-Gaussian, 25-Gaussian and the Swiss roll distributions, respectively. The generator is held fixed at real data plus unit-variance Gaussian noise. It is seen that discriminators trained with gradient penalty as well as spectral normalization have failed to capture the high moments of the data distribution. }
	\label{fig:fig1}	
\end{figure*}

To visualize the differences, we still use the synthetic datasets to test discriminators' performance. Figure \ref{fig:fig1} illustrates the value surfaces of the discriminators. 
It is obvious that discriminators trained with gradient penalty as well as spectral normalization have pathological value surfaces even when optimization has completed, and they have failed to capture the high moments of the data distributions and instead model very simple approximations to the optimal functions. In contrast, BC-GANs have successfully learned the higher moments of the data distributions, and the discriminator can distinguish the real distribution from the fake one much better.

\subsection{Convergence Measure}
One advantage of using Wasserstein distance as the metric over KL divergence is the meaningful loss. The Wasserstein distance $W(P_r, P_g)$ shows the property of convergence \cite{A8}. If it stops decreasing, then the training of the network can be terminated. This property is useful as one does not have to stare at the generated samples to figure out the failure modes.
To obtain the convergence measure in the proposed BC-GAN, a corresponding indicator of the training stage is introduced:
\begin{equation}\label{key15}
I_{GD}{\rm{ = }}\frac{1}{{||{\nabla _x}f(x)|{|_2}}}
\end{equation}

To prove that proposed indicator $I_{GD}$ is capable of convergence measure, Theorem 3 is introduced.

{\noindent \bf Theorem 3.} Let $P_r$ and $P_g$ be the distributions of real and generated images, $x$ is the image located in $P_r$ and $P_g$, and $f$ is the discriminator function, bounded by the $BC$ Conditions. $I_{GD}$ in Equation \ref{key15} is  proportional to $W(P_r, P_g)$.

\section{Experiments}
\subsection{Experiment setup}

In order to assess the performance of BC-GAN, image generation experiments are conducted on CIFAR-10 \cite{A13}, STL-10 \cite{A21} and CELEBA  \cite{A14}  datasets. Two widely-used GAN architectures, including the standard CNN and ResNet based CNN \cite{A8}, are applied for image generation task. For the architecture details, please see Appendix. Equations (\ref{key14}) and (\ref{key12}) are used as the loss metric of $D$ and $G$, respectively. $I_{GD}$ in Equation (\ref{key15}) acts as the role of measuring convergence. $m$ and $\beta$ in Equation (\ref{key14}) are set as 0.5 and 2, respectively. For optimization, the Adam \cite{A15} is utilized in all the experiments with $\alpha$=0.0002, $\beta_1=0$, $\beta_2=0.9$.  $D$ updates 5 times per $G$ update.  To keep it identical to previous GANs, we set the batch size as 64. 
Inception Score \cite{A16} and Fr\'{e}chet Inception Distance \cite{A21} are utilized for quantitative assessment of generated examples. 

Although Inception Score and  Fr\'{e}chet Inception Distance are widely used as an evaluation metric for GANs, Barratt \cite{A17} suggests that it should be more systematic and careful when evaluating and comparing generative models. Because inception score may not correlate well with the image quality strictly. Recently, Catherine \cite{A18} proposes a new method to evaluate the generative models, called skill rating. Skill rating evaluates models by carrying out tournaments between the discriminators and generators. For better evaluation, results assessed by skill rating is also presented.

\subsection{Results on Image Generation}

\begin{table*}[htp]
	\centering
	\begin{tabular}{c c c c c}
		\hline \hline
		\multirow{2}{*}{Method} & \multicolumn{2}{c}{CIFAR-10}  &\multicolumn{2}{c}{STL-10} \\ 
		&IS&FID 	&IS&FID \\
		\hline
		Real data & 11.24$\pm$.12 &7.8 &26.08$\pm$.26&7.9\\
		\hline
		{\bf -Standard CNN-} & \ &&&\\
		DCGAN  & 6.64$\pm$.14 & &7.84$\pm$.07&\\
		WGAN-GP  & 6.53$\pm$.08 &40.2 & 8.42$\pm$.13 &55.1\\
		SN-GAN  & 7.42$\pm$.06 & 29.3 &8.28$\pm$.09 &53.1 \\
		{\bf BC-GAN} & {\bf 7.48$\pm$.06} & {\bf28.9} &8.30 $\pm$.12& 54.5\\
		\hline
		{\bf -ResNet-}& &&& \\
		WGAN-GP  & 7.86$\pm$.13 \\
		SN-GAN  & 8.22$\pm$.05 &21.7 & 9.10$\pm$.04& 40.1\\
		{\bf BC-GAN} & {\bf 8.40$\pm$.10} & {\bf20.8}& {\bf9.15$\pm$.17}& {\bf39.9}\\
		\hline
		LR-GAN \cite{A19} & 7.17$\pm$.07&&&\\
		DFM \cite{A20} & 7.72$\pm$.13&&8.51$\pm$.12&\\
		Orthonormal\cite{A10} & 7.40$\pm$.04&29&8.56$\pm$.09&46.7\\
		\hline \hline
	\end{tabular} 
	\caption{IS and FID of unsupervised image generation on CIFAR-10 and STL-10. IS is the Inception Score, and FID represents Fr\'{e}chet Inception Distance. For IS, higher is better, while lower is better for FID.}
	\label{t1}
\end{table*}

\begin{figure*}[htp]
	\centering
	\begin{subfigure}[b]{.32\textwidth}
		\centering
		\includegraphics[width=1\linewidth]{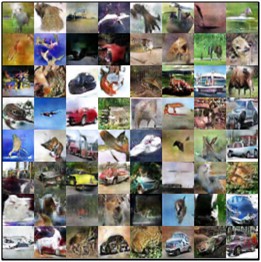}  
		\vspace*{-0.0cm}
		\caption{SN-GAN}
	\end{subfigure}
	\begin{subfigure}[b]{.32\textwidth}
		\centering
		\includegraphics[width=1\linewidth]{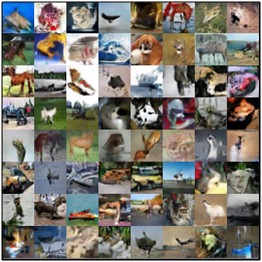}  
		\vspace*{-0.0cm}
		\caption{WGAN-GP}
	\end{subfigure}
	\begin{subfigure}[b]{.32\textwidth}
		\centering
		\includegraphics[width=1\linewidth]{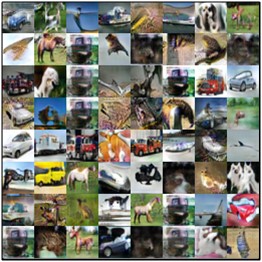}  
		\vspace*{-0.0cm}
		\caption{BC-GAN}
	\end{subfigure}
	\vspace*{-0.0cm}
	\caption{Image generation on CIFAR-10 dataset using (a) SN-GAN, (b) WGAN-GP and (c) BC-GAN.}
	\label{fig:fig2a}
	\begin{subfigure}[b]{.32\textwidth}
		\centering
		\includegraphics[width=1\linewidth]{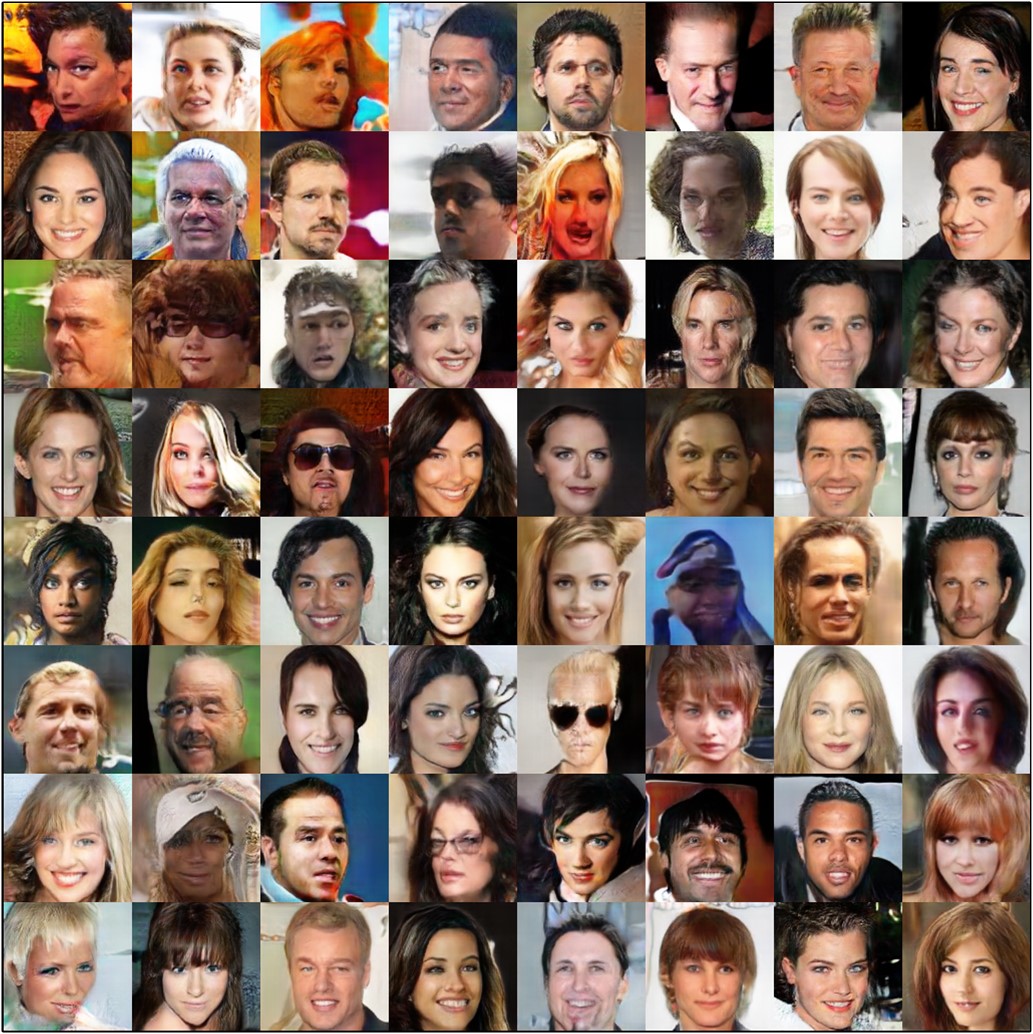}  
		\vspace*{-0.0cm}
		\caption{SN-GAN}
	\end{subfigure}
	\begin{subfigure}[b]{.32\textwidth}
		\centering
		\includegraphics[width=1\linewidth]{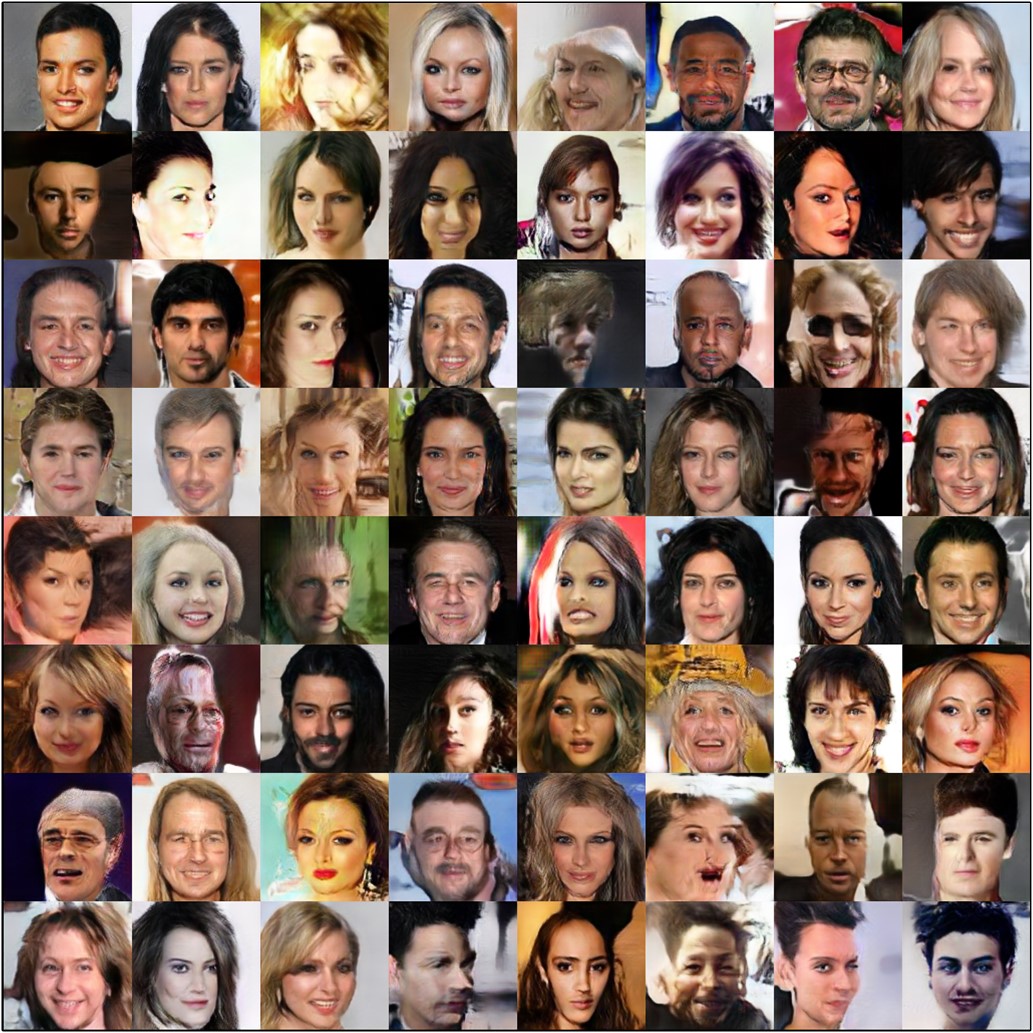}  
		\vspace*{-0.0cm}
		\caption{WGAN-GP}
	\end{subfigure}
	\begin{subfigure}[b]{.32\textwidth}
		\centering
		\includegraphics[width=1\linewidth]{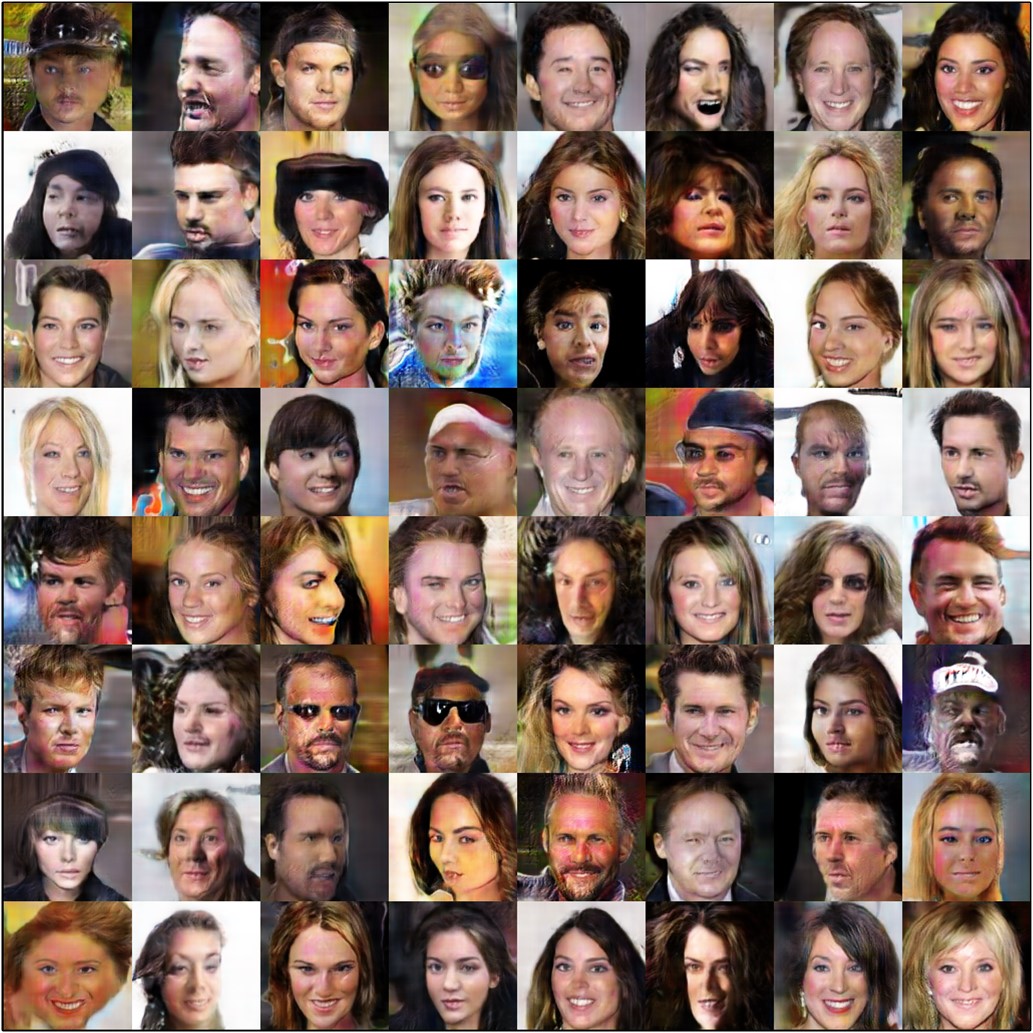}  
		\vspace*{-0.0cm}
		\caption{ BC-GAN}
	\end{subfigure}
	\vspace*{-0.0cm}
	\caption{Image generation on CELEBA dataset using (a) SN-GAN, (b) WGAN-GP and (c) BC-GAN.}
	\label{fig:fig2b}
	\vspace*{-0.5cm}
\end{figure*}

Image generation tasks are carried out on the CIFAR-10 and STL-10 datasets. Based on the ResNet based CNN architecture, we obtain the average inception score of 8.40 and 9.15 for image generation on CIFAR-10 and STL-10, respectively.
We compare our algorithm against multiple benchmark methods. In Table \ref{t1}, we show the Inception Score and Fr\'{e}chet Inception Distance of different methods with their corresponding optimal settings on CIFAR-10 and STL-10 datasets. As illustrated in Table \ref{t1}, BC-GAN has comparable performances with the state-of-the-art GANs.
We also conduct image generation on CELEBA \cite{A14} dataset. Examples of generated images are shown in Figure \ref{fig:fig2a} and \ref{fig:fig2b}.

\begin{figure}[h]
	\centering
	\includegraphics[width=0.5\linewidth]{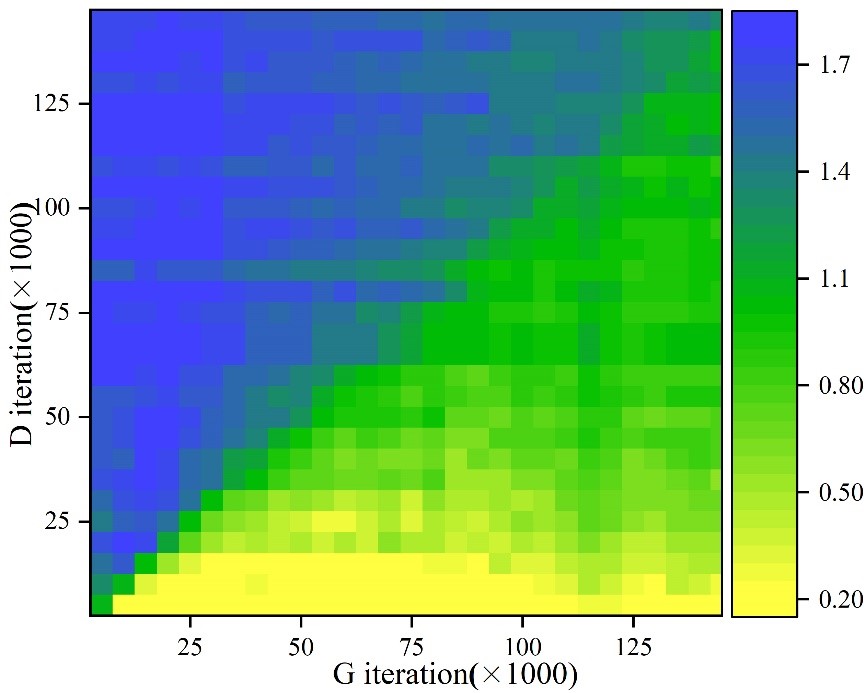}
	\caption{Matches between $D$ and $G$. Wasserstein distance is utilized to indicate the results instead of the win rate. With larger value of the Wassserstein distance, $D$ is more likely to distinguish the real images from the fake ones. Lower value of the Wasserstein distance indicates that $G$ is more likely to fool $D$}
	\label{fig:fig3}
\end{figure}

Skill rating \cite{A18} is recently introduced to judge the GAN model by matches between $G$ and $D$. To determine the outcome of a match between  $G$ and $D$,  $D$ judges two batches: one batch of samples from  $G$, and one batch of real data. Every sample \textit{x} that is not judged correctly by $D$ (e.g. $D$(\textit{x}) $>$0.5 for the generated data or $D(x)$ $<$0.5 for the real data) counts as a win for $G$ and is used to compute its win rate. Win rate tests the performance between $D$ and $G$ dynamically in the training process and judges whether $D$ or $G$ dominates, while the other stops updating. If $D$ dominates and $G$ stops updating, win rate for $G$ decreases dramatically. We make some modifications, because we use Wasserstein distance to determine the difference between fake  and real data instead of probability. As a result, we show the loss of $D$ instead of the win rate in Figure \ref{fig:fig3}. When $D$ in the latter iteration is used to distinguish the generated images in the early iteration from real images, it outputs a large loss, meaning that $D$ can easily distinguish the generated images (fake images) from real images. And the images generated in the latter iteration can also easily fool $D$ in the early iteration. Therefore, there is a healthy training, and the performance of $D$ and $G$ is continuously improved in the training process.

When applying KL divergence as the loss metric of $D$, the training of GANs suffers from the vaninshing gradient problem, \textit{i.e.}, zero gradient would back propagate to $G$, and the training would completely stop.  As a comparsion, Figure \ref{fig:fig3} shows a healthy training during the entire iterations, further indicating the effectiveness of BC-GANs.

\section{Analysis}

\subsection{ Bound $m$}

The parameter \textit{m} in Equation (\ref{key14}) represents the bound of $D$, and it actually controls the gradient $\partial$\textit{$L_{D}$}/$\partial$\textit{x}, where $L_D$ is the loss of $D$, $x$ is the image and $\partial$\textit{$L_{D}$}/$\partial$\textit{x} is the gradient backpropagated from $D$ to $G$, which indeed affects the training of $G$, and further influences the model performance. Explanation is as followed. The discriminator $f$ is a bounded function. Given enough iterations, \textit{f\textsubscript{x$\sim$Pr}}(\textit{x}) would always converge to $m$ and \textit{f\textsubscript{x$\sim$Pg}}(\textit{x}) would converge to $-m$. And considering that $f$ satisfies $k$-Lipschitz constraint, the following condition is satisfied:
\begin{equation}\label{key16}
||{f_{{x_r}\sim{{P_r}}}}({x_r}) - {f_{_{{x_g}\sim{{P_g}}}}}({x_g})|| \approx 2m \le k||{x_r} - {x_g}||
\end{equation}
\begin{equation}\label{key17}
\frac{{2m}}{{||{x_r} - {x_g}||}} \le k
\end{equation}

\begin{figure*}[htp]
	\centering
	\begin{subfigure}[b]{.3\textwidth}
		\centering
		\includegraphics[width=1\linewidth]{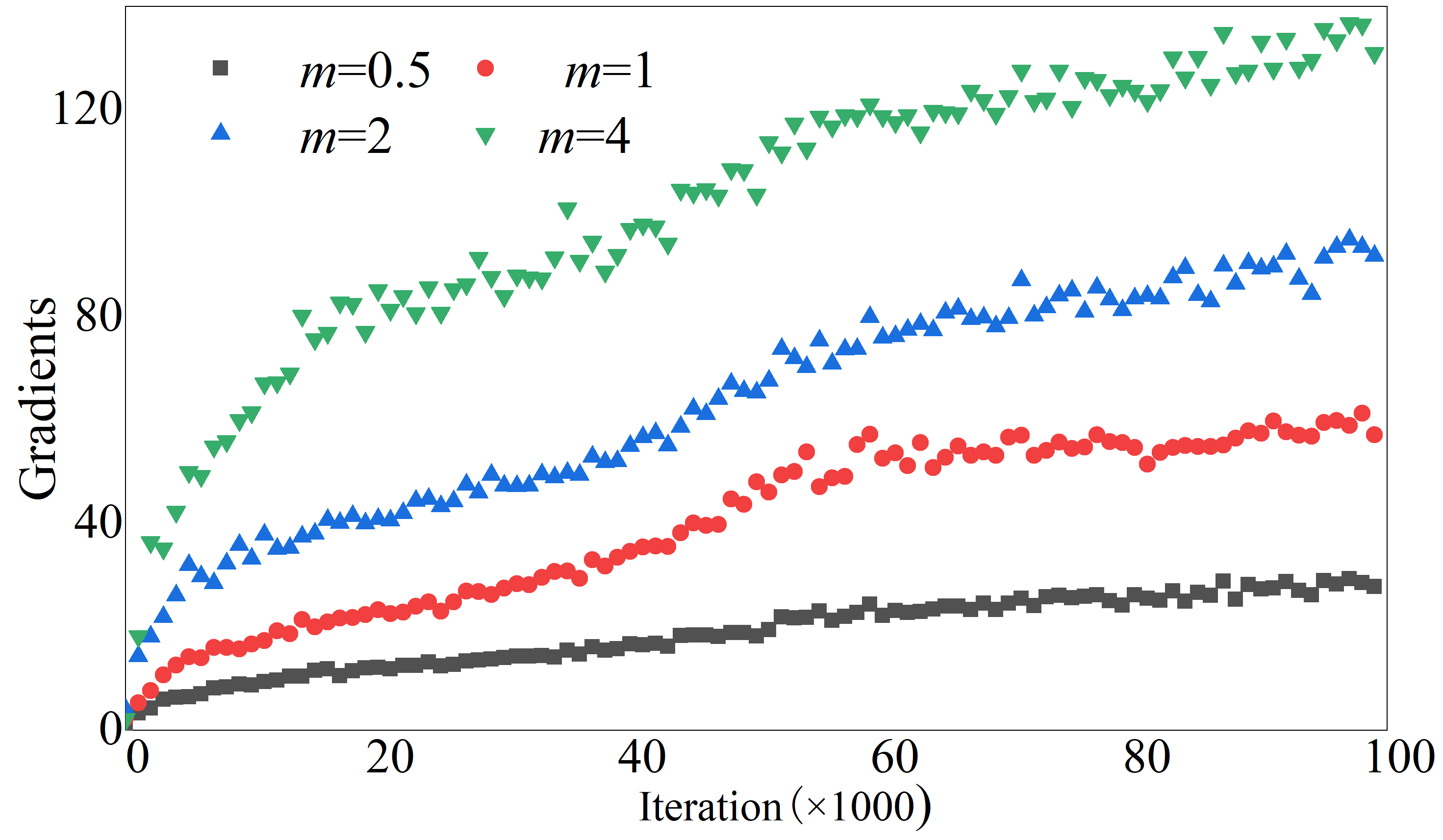}  
		\vspace*{-0.2cm}
		\caption{}
	\end{subfigure}	
	\begin{subfigure}[b]{.3\textwidth}
		\centering
		\includegraphics[width=1\linewidth]{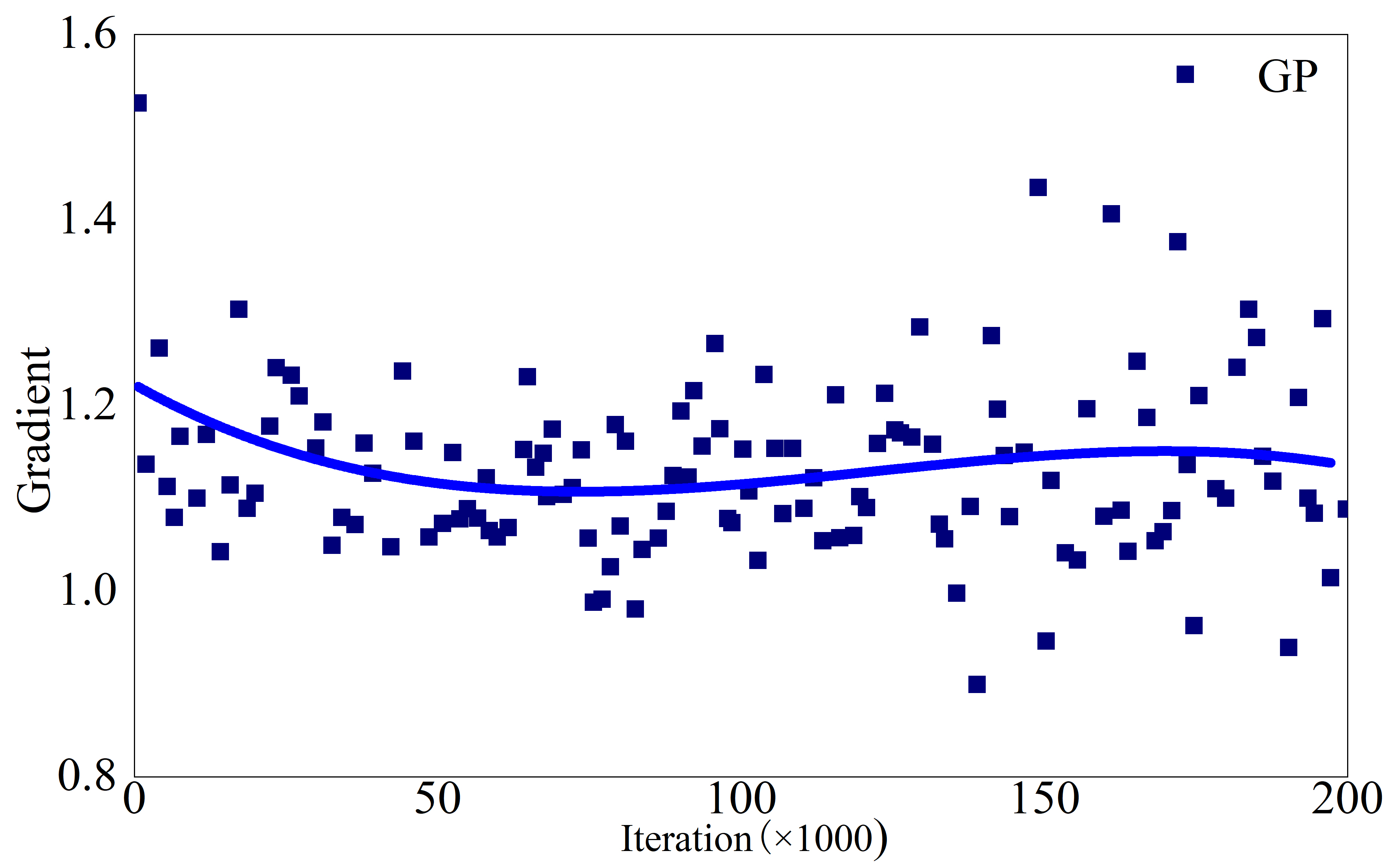}  
		\vspace*{-0.2cm}
		\caption{}
	\end{subfigure}
	\begin{subfigure}[b]{.3\textwidth}
		\centering
		\includegraphics[width=1\linewidth]{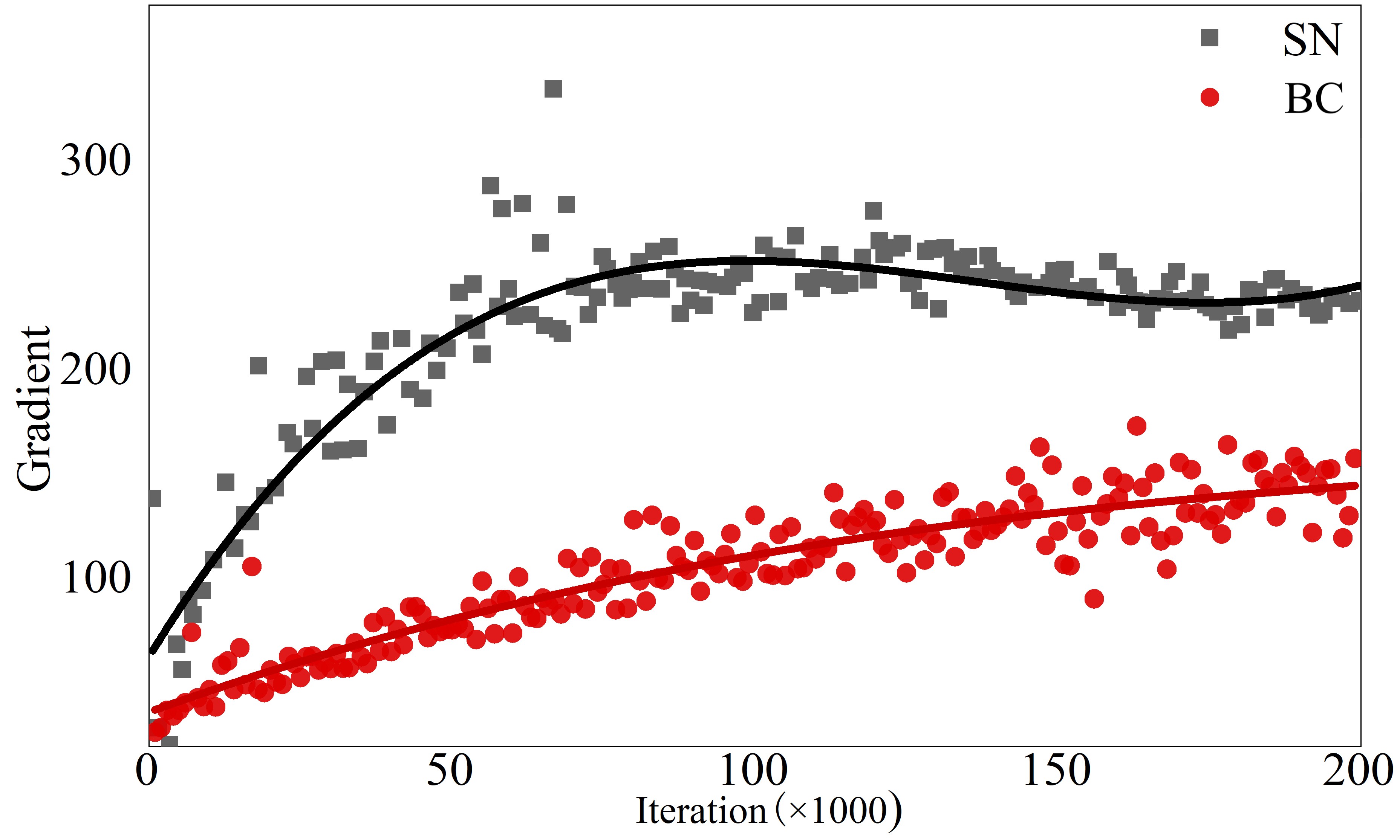}  
		\vspace*{-0.2cm}
		\caption{}
	\end{subfigure}
	\vspace*{-0.3cm}
	\caption{(a) variation of the gradient $\partial$\textit{L\textsubscript{D}}/$\partial$\textit{x} with iterations in BC-GAN.  Larger \textit{m} leads to higher gradients. (b)  variation of the gradient with iterations in WGAN-GP. (c) comparison of the gradient variation of SN-GAN and BC-GAN, where SN represents SN-GAN, and BC is BC-GAN.}
	\label{fig:fig4}
\end{figure*}

$k$ determines the upper bound of the gradient backpropagated from $D$ to $G$, and is directly proportional to $D$. Increasing $m$ enhances the upper bound of the gradients $\partial$\textit{L\textsubscript{D}}/$\partial$\textit{x}. This is verified by the experiment shown in Figure \ref{fig:fig4} (a). Moreover, the gradients are used to guide the training of the generator, and naturally affect the performance of the model.  Increasing $m$ from 0.5 to 2 leads to  decreased performance (Inception score drops from 8.40 to 7.56). Therefore,  properly controlling the gradient is  important for improving the performance of GAN models. And the bound $m$ provides such a mechanism for controlling the gradient. $m$ is recommended to be taken as 0.5 for image generation task on CIFAR-10. One possible explanation why a smaller $m$ (hence smaller gradients back-propagated) in the training leads to better performances is that the error surfaces are highly nonlinear, the backpropagation is a gradient descent and greedy algorithm, small gradients may help the optimization lead to a deeper local minimum or indeed the global minimum of the error surface.

We also monitor the variation of the gradient on WGAN-GP and SN-GAN. It's found that the behaviour of the gadient variation varys on different models. The gradient penalty term in WGAN-GP forces the gradient of the output of $D$ with respect to the input to be a fixed number. Therefore,  as shown in Figure \ref{fig:fig4} (b), the gradient is around 1 in the whole training process. 
For SN-GAN and our BC-GAN in Figure \ref{fig:fig4} (c), the variation of the gradient is similar. With training process going on, the gradient tends to increase until convergence is reached. The difference is that the amplitude of the gradient in SN-GAN is larger than that in BC-GAN. As mentioned above, the amplitude of the gradient indeed affects the training of the generator. However, SN-GAN provides no mechanism for controlling the gradient. While the bound $m$ in BC-GAN acts as the role of controlling the gradient. Thus, at least in this perspective, BC-GAN has a better performance control over SN-GAN.

\subsection{ Meaningful Training Stage Indicator $I_{GD}$}

\begin{figure*}[htp]
	\centering
	\begin{subfigure}[b]{.45\textwidth}
		\centering
		\includegraphics[width=1\linewidth]{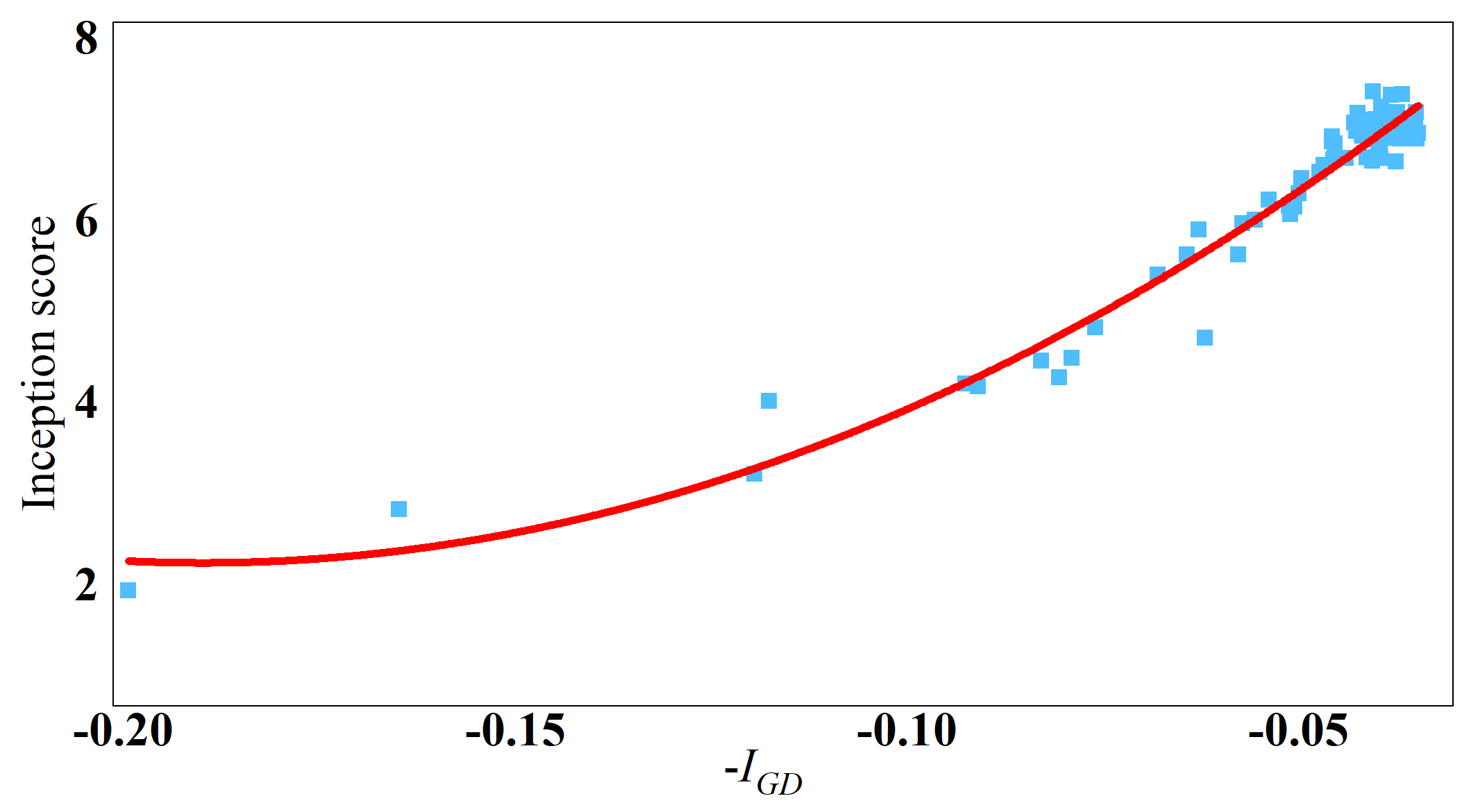}  
		\vspace*{-0.2cm}
		\caption{}
	\end{subfigure}	
	\begin{subfigure}[b]{.45\textwidth}
		\centering
		\includegraphics[width=1\linewidth]{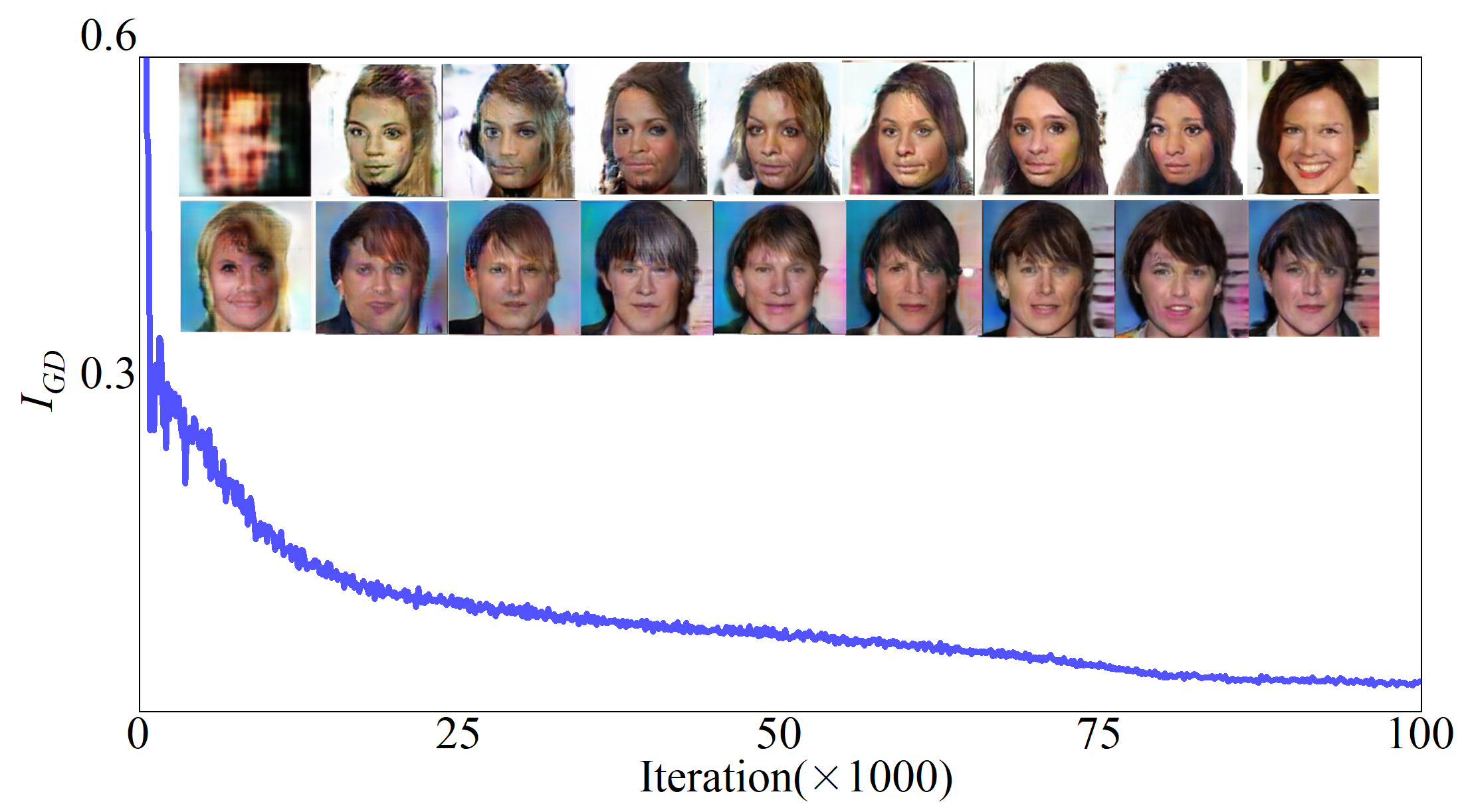}  
		\vspace*{-0.2cm}
		\caption{}
	\end{subfigure}
	\vspace*{-0.3cm}
	\caption{(a) correlation of -\textit{I\textsubscript{GD}} with inception score on CIFAR10. (b) variation of \textit{I\textsubscript{GD}} with iteration for the training on CELEBA database. \textit{I\textsubscript{GD}} correlates well with the image quality, indicating that \textit{I\textsubscript{GD}} can be regarded as the indicator of the training stage.}
	\label{fig:fig5}
	\vspace*{-0.0cm}
\end{figure*}

We introduce a new indicator $I_{GD}$ for monitoring the training stage. Figure \ref{fig:fig5}  (a) shows the correlation of$-I_{GD}$ with inception score during the training process. Because $I_{GD}$ decreases with the iteration, we use $-I_{GD}$ instead. As we can see, $-I_{GD}$ has a positive correlation with the inception score. As it is easier to visualize the correlation between $I_{GD}$ and image quality in higher resultion images, we perform image generation task on CELEBA \cite{A14} dataset and show the variation of $I_{GD}$ with iterations in  Figure \ref{fig:fig5} (b) . It's clearly seen that $I_{GD}$ correlates well with image quality during the training process.

\begin{figure}
	\centering
	\includegraphics[width=0.4\linewidth]{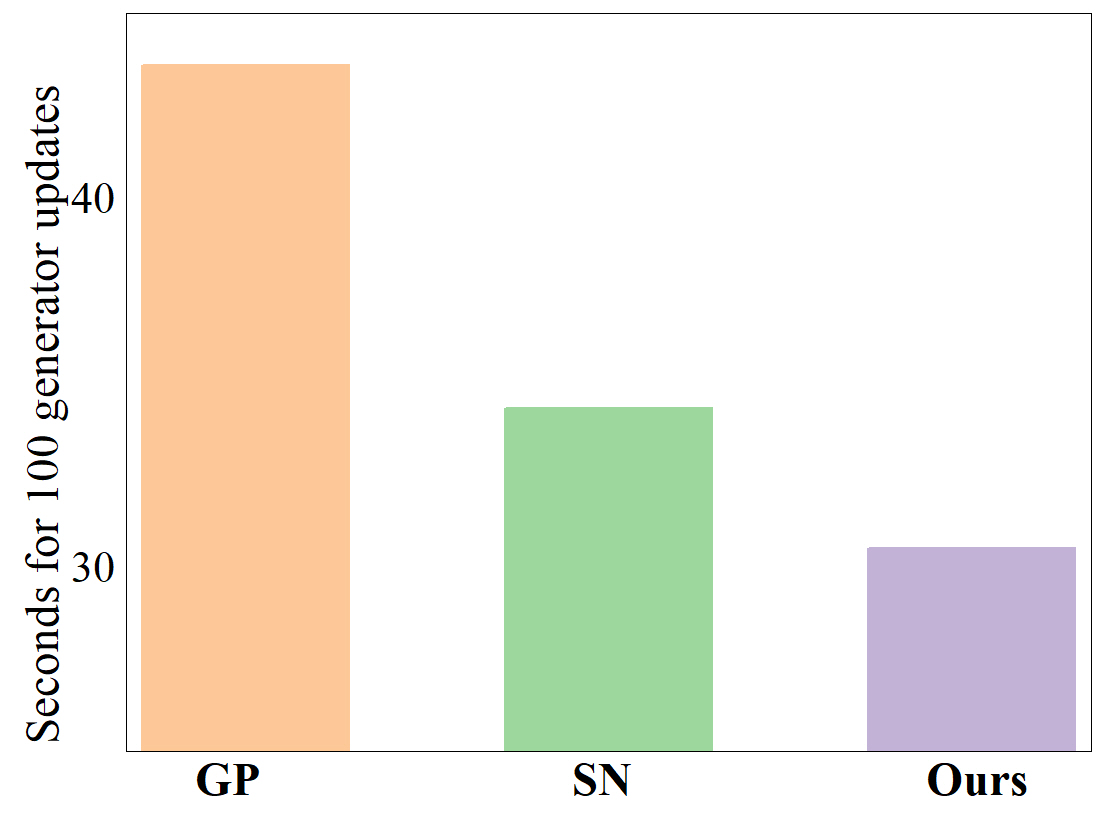}
	\caption{Computation time for 100 generator updates. GP for WGAN-GP and SN for SN-GAN. We use Standard CNN as the architecture. Tests are based on Nvidia 1080Ti.}
	\label{fig:fig6}
\end{figure}

\subsection{Training Time}		

It is worth noting that BC-GAN is computationally efficient. We list the computational time for 100 generator updates in Figure \ref{fig:fig6}. WGAN-GP requires more computational time because it needs to calculate the gradient of the gradient norm $\|$$\triangledown$\textsubscript{\textit{x}}$D$$\|$\textsubscript{2}, which needs one whole round of forward and backward propagation. And spectral normalization needs to calculate the largest singular value of the matrices in each layer. What is worse, for gradient penalty and spectral normalization, the extra computational costs increase with the increase of layers. As for BC-GAN, there is no matrix operation or gradient calculation in the backpropagation. As a result, it has lower computational cost.

\section{Concluding Remarks}

In this paper, we have introduced a new generative adversarial network training technique called BC-GAN which utilizes bounded discriminator to enforce Lipschitz constraint. In addition to provide theoretical background, we have also presented practical implementation procedures for training BC-GAN. Experiments on synthetical as well as real data show that the new BC-GAN performs better and has lower computational complexity than recent techniques such as spectral normalization GAN (SN-GAN) and Wasserstein GAN with gradient penalty (WGAN-GP). We have also introduced a new training convergence measure which correlates directly with the image quality of the generator output and can be conveniently used to monitor training progress and to decide when training is completed.


%
\section*{Conflict of Interest}
The authors declare that they have no conflict of interest.
We declare that we do not have any commercial or associative interest that represents a conflict of interest in connection with the work submitted

\bibliographystyle{spmpsci}      
\bibliography{mybibfile}   


\newpage

\appendix	
\appendixpage 
\section{Proofs}

Let $P_r$ and $P_g$ be the distributions of real images and generated images in $X$, a compact metric space. The discriminator function $f$ is constructed based on a neural network of the following form with input $x$:
\begin{equation} \label{key18}
f(x,\theta )=W^{L+1}a_L(W^L(a_{L-1}(\cdots a_1(W^1x)))))
\end{equation}	
where $\theta :=\{ W^1, W^2, ..., W^{L+1} \}$ is the learning parameter set, and $a_l$ is an element-wise non-linear activation function.  Spectral normalization is applied on $f$ to guarantee the Lipschitz constraint.  

{\noindent \bf Proposition 1} When using Wasserstein distance as the loss metric of $f$, the optimal solution to $f$  is unreachable.

{\noindent \bf Proof:}
The Corollary 1 in \cite{A8} has proven that the optimal discriminator $f^*$ has gradient norm 1 almost everywhere under $P_r$ and $P_g$ when using Wasserstein distance as the loss metric.

Suppose $x$ can be expressed as $[x_1, x_2, \cdots, x_n]$, and $W^TW$ has eigenvalues $\lambda= [\lambda_1, \lambda_2,\cdots, \lambda_n ]$:
\begin{equation}
\lambda_1  \geqslant  \lambda_2 \geqslant \cdots \geqslant  \lambda_n \geqslant 0
\end{equation}

The eigenvectors of $W^TW$ can be expressed as $V = [v_1, v_2, \cdots, v_n]$. Then, we have:
\begin{equation}
||Wx||^2 = x^TW^TWx = x^TV^T\lambda Vx
\end{equation}

Supposing the transformation $Vx= [y_1, y_2, \cdots, y_n]$, and using the relationship $V^TV= I$, we can have
\begin{equation}
\begin{split}
x^TV^T\lambda Vx &=\lambda_1 y_1^2 + \lambda_2 y_2^2 + \cdots + \lambda_n y_n^2\\
&\leqslant \lambda_1 (y_1^2 +  y_2^2 + \cdots + y_n^2)\\
&= \lambda_1 (x^TV^TVx) = \lambda_1 ||x||^2
\end{split}
\end{equation}

When spectral normalization is applied, $\lambda_1$ is normalized to 1. As a result:
\begin{equation}
||Wx||^2 \leqslant ||x||^2
\end{equation}

We can see that applying spectral normalization can guarantee $W$ satisfy the Lipschitz constraint. 
The discriminator function $f$ is implemented by covolutional neural networks, which is a combination of convolutional and non-linear operations (Equation (\ref{key18})). Therefore, the following inequality is applied to observe the bound on $||f||_{Lip}$ \cite{A10}:
\begin{equation}\label{keyA6}
||f||_{Lip} \leqslant \prod_{l=1}^{L+1} \sigma(W^l) \cdot \prod_{l=1}^{L+1} ||a_l||_{Lip}=1
\end{equation}
where $\sigma(W)$ is the spectral norm of $W$.

When applying Wasserstein distance as the loss metric, Corollary 1 in \cite{A8} has proven that the optimal solution to the Lipschitz constrained discriminator has gradient norm 1 almost everywhere under $P_r$ and $P_g$, which means $||f||_{Lip}$ needs to reach the upper bound of 1. However, if $||f||_{Lip}$ in Equation (\ref{keyA6})  needs to obtain the upper bound 1, the discriminator function becomes a linear function. Because the discriminator function is implemented by the combination of convolutional operation and non-linear operation. Taking the Relu function as a representation of the non-linear operation $a_l$, $a_l(x) = x (x > 0)$ or $a_l(x) =0 (x \leqslant 0)$. In another word, $||a_l(x)|| = ||x|| (x > 0)$, and $||a_l(x)|| = 0 < ||x|| (x \leqslant 0)$.
If the discriminator function needs to obtain the upper bound of the Lipschitz constraint, all the non-linear operations need to reach the upper bound as well: $||a_l(x)|| = ||x||$. Then, all the non-linear functions are linear functions, and the discriminator function turns to a linear function. Obviously, a linear discriminator is not the optimal solution. Therefore, with the existence of non-linear operation, applying spectral normalization prevents the discriminator functions from the optimal solution when applying Wasserstein distance as the loss metric.

{\noindent \bf Theorem 1.} Let $\Psi$ be the set of all $f: X  \rightarrow R $, where $f$ is a continuous  function. In addition, the number of points where $f$ is continuous but not differentiable is finite. Besides, if $f$ is differentiable at point $x$, its derivative is finite. Then, $f$ in $\Psi$ satisfies Lipschitz constraint.

{\noindent \bf Proof:}
(i) Considering that $f$ is derivable. According to Lagrange's Mean Value Theorem,
\begin{equation}
\frac{f(x_1)-f(x_2)}{x_1-x_2} = \frac{\partial f(x_0)}{\partial x_0} (x_0 \in [x_1, x_2])
\end{equation}

Because $\frac{\partial f}{\partial x}$ is finite:

\begin{equation}
\frac{f(x_1)-f(x_2)}{x_1-x_2} = \frac{\partial f(x_0)}{\partial x_0}\leqslant k
\end{equation}
where $k$ is finite.

Moreover, we have:
\begin{equation}
||f(x_1) - f(x_2)|| \leqslant k||x_1 -x_2||
\end{equation}
Then, $f$ satisfies Lipschitz constraint.

(ii) Considering that $f$ is not derivable. $f$ is a continuous function, then, there must be at least one point $x_0$, at which $f$ is continuous but not derivable. We only consider that there is only one such point. For multiple points, the conclusion is the same.
For any $x_1$ and $x_2$ ($x_1, x_2<x_0$ or $x_1, x_2>x_0$), $f$ should satisfy the following:
\begin{equation}
||f(x_1)-f(x_2)|| \leqslant k||x_1-x_2||
\end{equation}
because $f$ is continuous and derivable in $[x_1, x_2]$.

For $x_1$ and $x_2$ ($x_1<x_0<x_2$), we have
\begin{equation}
\begin{split}
||f(x_1)-f(x_2)||&=||f(x_1)-f(x_0)+f(x_0)-f(x_2)||\\
&\leqslant ||f(x_1)-f(x_0)||+||f(x_0)-f(x_2)||
\end{split}
\end{equation}

Because $f$ is continuous in $[x_1, x_0]$ and $[x_0, x_2]$, and derivable in $(x_1, x_0)$ and $(x_0, x_2)$, we can obtain:
\begin{equation}
||f(x_1)-f(x_0)||\leqslant k_1||x_1-x_0||
\end{equation} 
\begin{equation}
||f(x_0)-f(x_2)||\leqslant k_2||x_0-x_2||
\end{equation} 

Then, we can have:
\begin{equation}
||f(x_1)-f(x_2)||\leqslant k(||x_1-x_0||+||x_0-x_2||)
\end{equation} 
where $k=max(k_1, k_2)$. Considering the relationship that $x_1<x_0<x_2$, we can have:
\begin{equation} \label{keyA15}
||f(x_1)-f(x_2)||\leqslant k(||x_1-x_2||)
\end{equation} 

As we can see, even though $f$ is not derivable at $x_0$, for any $x_1$ and $x_2$, $f$ still satisfies: $||f(x_1)-f(x_2)||\leqslant k(||x_1-x_2||)$. 

To sum up, $f$ always satisfies Lipschitz constraint at the given conditions.

{\noindent \bf Theorem 2.}  Let $P_r$ and $P_g$ be the distributions of real images and generated images in $X$, a compact metric space. Let $\Omega$ be the set of all $f: X  \rightarrow R$, where $f$ is a continuous and bounded  function. And, the number of points where $f$ is continuous but not differentiable is finite. Besides, if $f$ is differentiable at point $x$, its derivative is finite. The set  $\Omega$ can be expressed as:
\begin{equation}
\Omega:{\rm{\{ }}f |  {\rm{ }}||f(x)|| \le m,{\rm{ if }}\frac{{\partial f(x)}}{{\partial x}}{\rm{exists, }}||\frac{{\partial f(x)}}{{\partial x}}|| < \infty {\rm{\} }}
\end{equation}
where $m$ represents the bound. Then, there must exist a $k$, and we have a computable $k \cdot W(P_r, P_g)$:
\begin{equation}\label{keyA17}
k \cdot W({P_r},{P_g}) = \mathop {\sup }\limits_{f \in \Omega } \mathop E\limits_{x \sim {P_r}} [f(x)] - \mathop E\limits_{x \sim {P_g}} [f(x)]
\end{equation}
where $W(P_r, P_g)$ represents the Wasserstein distance \cite{A1,A4} between $P_r$ and $P_g$.

{\noindent \bf Proof:}
According to Theorem 1, for $f$ in $\Omega$, there exists a $k$ to satisfy Equation (\ref{keyA15}). Then, $\Omega$ is the set, which contains all the $k$-Lipschitz constrained functions $f$. Kantorovich-Rubinstein duality \cite{A1,A4} tell us that the supremum over all the functions in $\Omega$ is $k\cdot W(P_r, P_g)$. As a result, we can obtain Equation (\ref{keyA17}).
To guarantee the boundedness and computability of $k\cdot W(P_r, P_g)$, $f$ is supposed to be a bounded function. Becasue, even though $k$ in Theorem 1 is a finite number, it can be super large $k \rightarrow \infty$, leading to the incomputability  of $k\cdot W(P_r, P_g)$. Enforcing $f$ to be a bounded function can ensure the boundedness and computability of $k\cdot W(P_r, P_g)$:
\begin{equation}
k \cdot W({P_r},{P_g}) = \mathop {\sup }\limits_{f \in \Omega } \mathop E\limits_{x \sim {P_r}} [f(x)] - \mathop E\limits_{x \sim {P_g}} [f(x)] \le 2m
\end{equation}

{\noindent \bf Theorem 3.} Let $P_r$ and $P_g$ be the distributions of real and generated images, $x$ is the image located in $P_r$ and $P_g$, and $f$ is the discriminator function, bounded by the $BC$ Conditions. $I_{GD}$ in Equation \ref{key15} is  proportional to $W(P_r, P_g)$.

{\noindent \bf Proof:}
$f$ is bounded by the $BC$ conditions. Given enough iterations, $f_{x \sim P_r}(x)$ would always converge to $m$ and $f_{x \sim P_g}(x)$ would converge to $-m$. As a result, $k \cdot W(P_r, P_g)$  will always converge to $2m$:
\begin{equation}
k \cdot W({P_r},{P_g}) = \mathop {\sup }\limits_{f \in \Omega } \mathop E\limits_{x \sim {P_r}} [f(x)] - \mathop E\limits_{x \sim {P_g}} [f(x)] \approx 2m
\end{equation}

It is clear that $W(P_r, P_g)$ is proportional to $E\left [ \left \| x_r - x_g \right \| \right ]$, because both of them evaluate the difference between $P_r$ and $P_g$.  Then, we can use the following term $GD$ to estimate $W(P_r, P_g)$:
\begin{equation}
GD = \frac{{\lVert{f_{{x_r}\sim{{P_r}}}}({x_r}) - {f_{_{{x_g}\sim{{P_g}}}}}({x_g})\lVert}}{{||{x_r} - {x_g}||}}
\end{equation}
where $x_r$, $x_g$ are the real image and generated image, respectively.
As expressed above, the term $ || f_{x_r \sim p_r}(x_r)-f_{x_g \sim p_g}(x_g) ||$
would always converge to $2m$, and $W(P_r, P_g)$ is proportional to $E\left [ \left \| x_r - x_g \right \| \right ]$.
Therefore, $GD$ is inversely related to $W(P_r, P_g)$ , and the reciprocal of $GD$ can be used to roughly estimate $W(P_r, P_g)$.

According to Lagrange's Mean Value Theorem,
\begin{equation}
GD = \frac{{||{f_{{x_r}\sim{{P_r}}}}({x_r}) - {f_{_{{x_g}\sim{{P_g}}}}}({x_g})||}}{{||{x_r} - {x_g}||}}{\rm{ = }}||{\nabla _x}f(x)|{|_2}
\end{equation}
where $x \in \left [ x_g, x_r \right ]$. For the convenience of calculation, $x$ is taken as $x = \alpha \cdot x_r + (1- \alpha) \cdot x_g$, and $\alpha \in$[0, 1]. Then, $\left \| \triangledown_xf(x) \right \|_2$  is  inversely related to $W(P_r, P_g)$. Finally, $I_{GD}$ is proportional to $W(P_r, P_g)$.

\section{Architecture}
Discriminator in the toy model is listed in Table \ref{t2}. Standard CNN architectures for CIFAR-10 and STL-10 are listed in Table \ref{t3} and \ref{t4}. ResNet based CNN architectures for CIFAR10 and STL-10 are listed in Table \ref{t5} and \ref{t6}. Architectures for image generation on CELEBA dataset are listed in Table \ref{t7} and \ref{t8}.

\begin{table*}[h]
	\centering
	\begin{tabular}{c}
		\hline \hline
		Input points : $x \in R^2$\\ 
		\hline
		Dense, Relu $\rightarrow 512 \times 2$ \\
		\hline
		Dense, Relu $\rightarrow 512 \times 2$ \\
		\hline
		Dense, Relu $\rightarrow 512 \times 2$ \\
		\hline
		Dense$\rightarrow 1$\\
		\hline \hline
	\end{tabular} 
	\caption{Discriminator in the toy model}
	\label{t2}
\end{table*}

\begin{table*}[h]
	\centering
	\begin{tabular}{c}
		\hline \hline
		Latent vector : $z \in R^{128} \sim N(0, 1)$\\ 
		\hline
		Dense, BN, Relu $\rightarrow 4 \times 4 \times 512$ \\
		\hline
		$5 \times 5$, stride=2, Deconv, BN, Relu $\rightarrow 8 \times 8 \times 256$ \\
		\hline
		$5 \times 5$, stride=2, Deconv, BN, Relu $\rightarrow 16 \times 16 \times 128$ \\
		\hline
		$5 \times 5$, stride=2, Deconv, BN, Relu $\rightarrow 32 \times 32 \times 64$ \\
		\hline
		$3 \times 3$, stride=1, Conv, Tanh $\rightarrow 32 \times 32 \times 3$ \\
		\hline \hline
	\end{tabular} 
	\caption{Generator of standard CNN architectures for CIFAR-10 and STL-10.}
	\label{t3}
\end{table*}

\begin{table*}[h]
	\centering
	\begin{tabular}{c}
		\hline \hline
		Input RGB image : $x \in R^{32 \times 32 \times 3}$\\ 
		\hline
		$3 \times 3$, stride=1, Conv, Leaky-Relu $\rightarrow 32 \times 32 \times 64$ \\
		\hline
		$5 \times 5$, stride=2, Conv, Leaky-Relu $\rightarrow 16 \times 16 \times 128$ \\
		\hline
		$5 \times 5$, stride=2, Conv, Leaky-Relu $\rightarrow 8 \times 8 \times 256$ \\
		\hline
		$5 \times 5$, stride=2, Conv, Tanh $\rightarrow 4 \times 4 \times 512$ \\
		\hline
		Dense $\rightarrow 1$\\
		\hline \hline
	\end{tabular} 
	\caption{Discriminator of standard CNN architectures for CIFAR-10 and STL-10.}
	\label{t4}
\end{table*}

\begin{table*}[h]
	\centering
	\begin{tabular}{c}
		\hline \hline
		Latent vector : $z \in R^{128} \sim N(0, 1)$\\ 
		\hline
		Dense $\rightarrow 4 \times 4 \times 128$ \\
		\hline
		ResBlock up $\rightarrow 8 \times 8 \times 128$ \\
		\hline
		ResBlock up $\rightarrow 16 \times 16 \times 128$ \\
		\hline
		ResBlock up $\rightarrow 32 \times 32 \times 128$ \\
		\hline
		BN, Relu $\rightarrow 32 \times 32 \times 128$ \\
		\hline
		$3 \times 3$, stride =1, Conv, Tanh $\rightarrow 32 \times 32 \times 3$ \\
		\hline \hline
	\end{tabular} 
	\caption{Generator of ResNet based CNN architectures for CIFAR10 and STL-10}
	\label{t5}
\end{table*}

\begin{table*}[h]
	\centering
	\begin{tabular}{c}
		\hline \hline
		Input RGB image : $x \in R^{32 \times 32 \times 3}$\\ 
		\hline
		$3 \times 3$, stride=1, Conv $\rightarrow 32 \times 32 \times 64$ \\
		\hline
		ResBlock down $\rightarrow 16 \times 16 \times 128$ \\
		\hline
		ResBlock down $\rightarrow 8 \times 8 \times 128$ \\
		\hline
		ResBlock down $\rightarrow 4 \times 4 \times 128$ \\
		\hline
		Dense $\rightarrow 1$\\
		\hline \hline
	\end{tabular} 
	\caption{Discriminator of ResNet based CNN architectures for CIFAR10 and STL-10}
	\label{t6}
\end{table*}

\begin{table*}[h]
	\centering
	\begin{tabular}{c}
		\hline \hline
		Latent vector : $z \in R^{128} \sim N(0, 1)$\\ 
		\hline
		Dense, BN, Relu $\rightarrow 4 \times 4 \times 512$ \\
		\hline
		Upsample $\rightarrow 8 \times 8 \times 512$\\
		\hline
		$3 \times 3$, stride=1, Conv, BN, Relu $\rightarrow 8 \times 8 \times 256$ \\
		\hline
		Upsample $\rightarrow 16 \times 16 \times 256$\\
		\hline
		$3 \times 3$, stride=1, Conv, BN, Relu $\rightarrow 16 \times 16 \times 128$ \\
		\hline
		Upsample $\rightarrow 32 \times 32 \times 128$\\
		\hline
		$3 \times 3$, stride=1, Conv, BN, Relu $\rightarrow 32 \times 32 \times 64$ \\
		\hline
		Upsample $\rightarrow 64 \times 64 \times 64$\\
		\hline
		$3 \times 3$, stride=1, Conv, BN, Relu $\rightarrow 64\times 64 \times 32$ \\
		\hline
		Upsample $\rightarrow 128 \times 128 \times 32$\\
		\hline
		$3 \times 3$, stride=1, Conv, BN, Relu $\rightarrow 128 \times 128 \times 32$\\
		\hline
		$3 \times 3$, stride =1, Conv, Tanh $\rightarrow 128 \times 128 \times 3$ \\
		\hline \hline
	\end{tabular} 
	\caption{Generator architecture for image generation on CELEBA dataset.}
	\label{t7}
\end{table*}

\begin{table*}[h]
	\centering
	\begin{tabular}{c}
		\hline \hline
		Input RGB image : $x \in R^{128 \times 128 \times 3}$\\ 
		\hline
		$3 \times 3$, stride=1, Conv, Leaky-Relu $\rightarrow 128 \times 128 \times 64$ \\
		\hline
		Downsample $\rightarrow 64 \times 64 \times 64$ \\
		\hline
		$3 \times 3$, stride=1, Conv, Leaky-Relu $\rightarrow 64 \times 64 \times 128$ \\
		\hline
		Downsample $\rightarrow 32 \times 32 \times 128$ \\
		\hline
		$3 \times 3$, stride=1, Conv, Leaky-Relu $\rightarrow 32 \times 32 \times 256$ \\
		\hline
		Downsample $\rightarrow 16 \times 16 \times 256$ \\
		\hline
		$3 \times 3$, stride=1, Conv, Leaky-Relu $\rightarrow 16 \times 16 \times 512$ \\
		\hline
		Downsample $\rightarrow 8 \times 8 \times 512$ \\
		\hline
		$3 \times 3$, stride=1, Conv, Leaky-Relu $\rightarrow 8 \times 8 \times 512$ \\
		\hline
		Downsample $\rightarrow 4 \times 4 \times 512$ \\
		\hline
		Dense $\rightarrow 1$\\
		\hline \hline
	\end{tabular} 
	\caption{Discriminator architecture for image generation on CELEBA dataset.}
	\label{t8}
\end{table*}

\end{document}